\documentclass[letterpaper, 10 pt, conference]{ieeeconf}  

\IEEEoverridecommandlockouts                              

\usepackage[dvipsnames,table]{xcolor}

\newcommand{\ST}[1]{\color{orange}}

\newcommand{\joon}[1]{{\color{black}#1}}
\newcommand{\rev}[1]{{\color{black}#1}}
\newcommand{\corr}[1]{{\color{black}#1}}

\usepackage{float}
\usepackage{pifont}
\usepackage{graphics}
\usepackage[pdftex]{graphicx}
\usepackage[tight,footnotesize]{subfigure}
\usepackage{multirow}
\usepackage[cmex10]{amsmath}
\usepackage{amssymb}
\usepackage{amsfonts}
\usepackage{algorithm}
\usepackage{booktabs}
\usepackage{textcomp}
\usepackage{url}
\usepackage{color}
\usepackage{siunitx}
\usepackage{soul}

\usepackage{threeparttable} 
\usepackage{threeparttablex} 
\usepackage{wrapfig}
\usepackage{babel}
\usepackage{placeins}
\usepackage{float}
\usepackage{algorithm}
\usepackage{algpseudocode}
\usepackage{makecell}

\usepackage{footnote}
\usepackage{tabularray}
\usepackage{stfloats}

\usepackage[colorlinks=magenta, linkcolor=magenta, citecolor=magenta, urlcolor=magenta]{hyperref}

\usepackage{cleveref}

\definecolor{White}{rgb}{1.,0.,1.}
\definecolor{first}{rgb}{.8,.0,.0}
\definecolor{second}{rgb}{.0,.6,.0}
\definecolor{third}{rgb}{.0,.0,.8}

\definecolor{ceiling}{RGB}{214,  38, 40}
\definecolor{floor}{RGB}{43, 160, 4}
\definecolor{wall}{RGB}{158, 216, 229}
\definecolor{window}{RGB}{114, 158, 206}
\definecolor{chair}{RGB}{204, 204, 91}
\definecolor{bed}{RGB}{255, 186, 119}
\definecolor{sofa}{RGB}{147, 102, 188}
\definecolor{table}{RGB}{30, 119, 181}
\definecolor{tvs}{RGB}{160, 188, 33}
\definecolor{furniture}{RGB}{255, 127, 12}
\definecolor{objects}{RGB}{196, 175, 214}

\definecolor{car}{rgb}{0.39215686, 0.58823529, 0.96078431}
\definecolor{bicycle}{rgb}{0.39215686, 0.90196078, 0.96078431}
\definecolor{motorcycle}{rgb}{0.11764706, 0.23529412, 0.58823529}
\definecolor{truck}{rgb}{0.31372549, 0.11764706, 0.70588235}
\definecolor{othervehicle}{rgb}{0.39215686, 0.31372549, 0.98039216}
\definecolor{person}{rgb}{1.        , 0.11764706, 0.11764706}
\definecolor{bicyclist}{rgb}{1.        , 0.15686275, 0.78431373}
\definecolor{motorcyclist}{rgb}{0.58823529, 0.11764706, 0.35294118}
\definecolor{road}{rgb}{1.        , 0.        , 1.        }
\definecolor{parking}{rgb}{1.        , 0.58823529, 1.        }
\definecolor{sidewalk}{rgb}{0.29411765, 0.        , 0.29411765}
\definecolor{otherground}{rgb}{0.68627451, 0.        , 0.29411765}
\definecolor{building}{rgb}{1.        , 0.78431373, 0.        }
\definecolor{fence}{rgb}{1.        , 0.47058824, 0.19607843}
\definecolor{vegetation}{rgb}{0.        , 0.68627451, 0.        }
\definecolor{trunk}{rgb}{0.52941176, 0.23529412, 0.        }
\definecolor{terrain}{rgb}{0.58823529, 0.94117647, 0.31372549}
\definecolor{pole}{rgb}{1.        , 0.94117647, 0.58823529}
\definecolor{trafficsign}{rgb}{1.        , 0.        , 0.        }
\definecolor{otherstructure}{rgb}{0.98039215, 0.58823529, 0.}
\definecolor{otherobject}{rgb}{0.19607843, 1.        , 1.        }

\makeatletter
\newcommand{\car@semkitfreq}{3.92}
\newcommand{\bicycle@semkitfreq}{0.03}
\newcommand{\motorcycle@semkitfreq}{0.03}
\newcommand{\truck@semkitfreq}{0.16}
\newcommand{\othervehicle@semkitfreq}{0.20}
\newcommand{\person@semkitfreq}{0.07}
\newcommand{\bicyclist@semkitfreq}{0.07}
\newcommand{\motorcyclist@semkitfreq}{0.05}
\newcommand{\road@semkitfreq}{15.30}
\newcommand{\parking@semkitfreq}{1.12}
\newcommand{\sidewalk@semkitfreq}{11.13}
\newcommand{\otherground@semkitfreq}{0.56}
\newcommand{\building@semkitfreq}{14.1}
\newcommand{\fence@semkitfreq}{3.90}
\newcommand{\vegetation@semkitfreq}{39.3}
\newcommand{\trunk@semkitfreq}{0.51}
\newcommand{\terrain@semkitfreq}{9.17}
\newcommand{\pole@semkitfreq}{0.29}
\newcommand{\trafficsign@semkitfreq}{0.08}
\newcommand{\semkitfreq}[1]{{\csname #1@semkitfreq\endcsname}}

\newcommand{\car@sscbkitfreq}{2.85}
\newcommand{\bicycle@sscbkitfreq}{0.01}
\newcommand{\motorcycle@sscbkitfreq}{0.01}
\newcommand{\truck@sscbkitfreq}{0.16}
\newcommand{\othervehicle@sscbkitfreq}{5.75}
\newcommand{\person@sscbkitfreq}{0.02}
\newcommand{\road@sscbkitfreq}{14.98}
\newcommand{\parking@sscbkitfreq}{2.31}
\newcommand{\sidewalk@sscbkitfreq}{6.43}
\newcommand{\otherground@sscbkitfreq}{2.05}
\newcommand{\building@sscbkitfreq}{15.67}
\newcommand{\fence@sscbkitfreq}{0.96}
\newcommand{\vegetation@sscbkitfreq}{41.99}
\newcommand{\terrain@sscbkitfreq}{7.10}
\newcommand{\pole@sscbkitfreq}{0.22}
\newcommand{\trafficsign@sscbkitfreq}{0.06}
\newcommand{\otherstructure@sscbkitfreq}{4.33}
\newcommand{\otherobject@sscbkitfreq}{0.28}
\newcommand{\sscbkitfreq}[1]{{\csname #1@sscbkitfreq\endcsname}}

\newcommand{\Fref}[1]{Fig.~\ref{#1}}
\newcommand{\Tref}[1]{Table~\ref{#1}}
\newcommand{\Eref}[1]{Eq.~(\ref{#1})}
\newcommand{\Sref}[1]{Sec.~\ref{#1}}

\usepackage{multirow}
\usepackage{wrapfig}
\usepackage{colortbl}
\usepackage[normalem]{ulem}

\newcolumntype{g}{>{\columncolor{CuGray}}c}
\newcolumntype{z}{>{\columncolor{CuGray}}l}

\renewcommand{\paragraph}[1]{\noindent\textbf{#1.}\,\,}

\usepackage{xspace}

\def\onedot{.\@\xspace}
\def\eg{\emph{e.g}\onedot} 
\def\ie{\emph{i.e}\onedot}

\newcommand{\be}{\begin{eqnarray}}
\newcommand{\ee}{\end{eqnarray}}
\newcommand{\bee}{\begin{eqnarray*}}
\newcommand{\eee}{\end{eqnarray*}}

\newcommand{\matrixb}{\left[ \begin{array}}
\newcommand{\matrixe}{\end{array} \right]}

\title{\LARGE \bf
VPOcc: Exploiting Vanishing Point for 3D Semantic \\ Occupancy Prediction
}

\author{Junsu Kim$^{1}$,  Junhee Lee$^{1}$,  Ukcheol Shin$^{2}$, Jean Oh$^{2}$ and Kyungdon Joo$^{1,\dag}$
\thanks{This work was supported by Institute of Information \& communications Technology Planning \& Evaluation~(IITP) grant funded by the Korea government~(MSIT) (No.RS-2020-II201336, Artificial Intelligence Graduate School Program~(UNIST)) and the National Research Foundation of Korea~(NRF) grant funded by the Korea government~(MSIT) (No.RS-2024-00457065 and No.RS-2025-02216916).}
\thanks{
$^{1}$Junsu Kim, Junhee Lee and Kyungdon Joo are with the Artificial Intelligence Graduate School, UNIST, Ulsan, South Korea. 
{\tt\small \{joonsu0109, junhee98, kyungdon\}@unist.ac.kr}
}
\thanks{
$^{2}$Ukcheol Shin and Jean Oh are with Robotics Institute, Carnegie Mellon University, Pittsburgh, Pennsylvania, 15217, United States. 
{\tt\small \{ushin, hyaejino\}@andrew.cmu.edu}
}
\thanks{$^{\dag}$Corresponding author}
}

\begin{document}
\maketitle

\thispagestyle{empty}
\pagestyle{empty}

\begin{abstract}
Understanding 3D scenes semantically and spatially is crucial for the safe navigation of robots and autonomous vehicles, aiding obstacle avoidance and accurate trajectory planning.
Camera-based 3D semantic occupancy prediction, which infers complete voxel grids from 2D images, is gaining importance in robot vision for its resource efficiency compared to 3D sensors.
However, this task inherently suffers from a 2D–3D discrepancy, where objects of the same size in 3D space appear at different scales in a 2D image depending on their distance from the camera due to perspective projection.
To tackle this issue, we propose a novel framework called \texttt{VPOcc} that leverages a vanishing point (VP) to mitigate the 2D-3D discrepancy at both the pixel and feature levels.
As a pixel-level solution, we introduce a VPZoomer module, which warps images by counteracting the perspective effect using a VP-based homography transformation.
In addition, as a feature-level solution, we propose a VP-guided cross-attention (VPCA) module that performs perspective-aware feature aggregation, utilizing 2D image features that are more suitable for 3D space.
Lastly, we integrate two feature volumes extracted from the original and warped images to compensate for each other through a spatial volume fusion (SVF) module.
By effectively incorporating VP into the network, our framework achieves improvements in both IoU and mIoU metrics on SemanticKITTI and SSCBench-KITTI360 datasets.
Additional details are available at \textcolor{magenta}{\url{https://vision3d-lab.github.io/vpocc/}}.
\end{abstract}

\begin{figure}[t]
    \centering
    \includegraphics[width=0.99\linewidth]{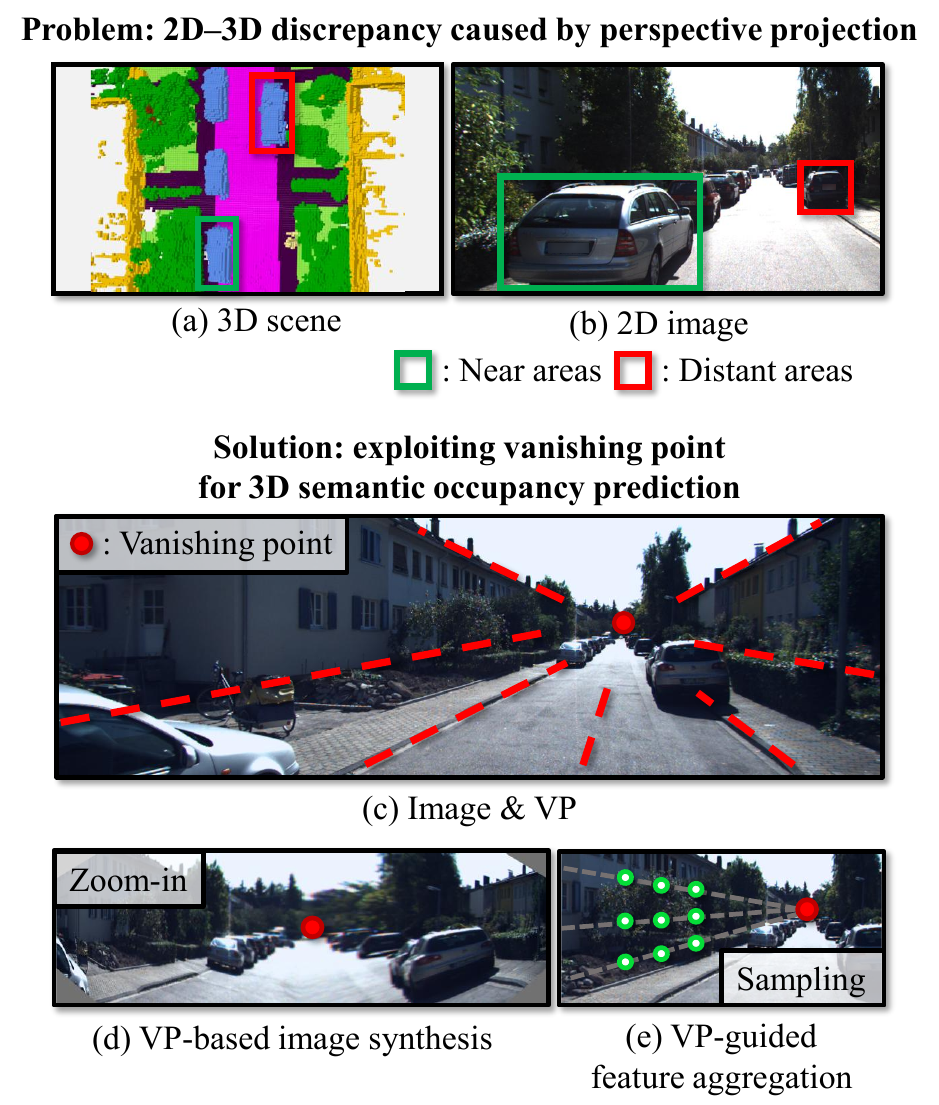}
    \caption{%
        \textbf{Overview of the proposed \texttt{VPOcc}}. 
        \emph{Problem (\corr{top}):} (a)~3D objects of similar size (\eg, \textcolor{ForestGreen}{left green box} and \textcolor{red}{right red box}) appear differently in (b) 2D image due to perspective projection, depending on their distance from the camera.
        \emph{Solution (\corr{bottom}):}
        We propose a framework that leverages a vanishing point (VP) to incorporate this discrepancy caused by perspective projection into the network and thereby utilize 2D image features in a 3D-aware manner.
        (c)~Given an input image and the VP, we perform (d)~VP-based image synthesis to zoom in, and (e)~VP-guided point sampling for perspective-aware feature aggregation.
    }
    \label{fig:overview}
    \vspace{4mm}
\end{figure}
\section{Introduction}  \label{sec:intro}
Comprehensive 3D scene understanding is a crucial challenge for robotic systems and intelligent agents, enabling accurate and effective path‑planning and navigation~\cite{wang2021learning, crespo2020semantic}.
Camera-based 3D semantic occupancy prediction, which estimates complete 3D scenes as voxel grids at the semantic level using only RGB cameras, is gaining attention due to its affordability compared to 3D sensors (\eg, LiDAR).
However, camera-based methods suffer from several problems, such as missing metric-scale depth~\cite{cao2022monoscene}, feature dimension mismatch~\cite{li2023voxformer, huang2023tri, yao2023ndc}, and occlusion~\cite{zhang2023occformer, jiang2024symphonize}.
We consider most of these issues to stem from the fundamental limitations inherent in the perspective projection of the camera.

The 2D image is generated by projecting a 3D scene onto the image plane via perspective projection, which inherently introduces 2D-3D discrepancy depending on the distance from the camera~\cite{hartley2003multiple}.
As illustrated in \Fref{fig:overview}-(a), the near object~(\textcolor{ForestGreen}{left green box}) and the distant object~(\textcolor{red}{right red box}), though similarly sized in 3D space, are projected at different scales in the 2D image.
We break down this problem into two perspectives: pixel and feature levels.
The pixel-level discrepancy indicates a \textit{discrepancy in the number of pixels} between near and distant areas in the 2D image.
Additionally, it causes the feature-level discrepancy, meaning an \textit{imbalance in the feature granularity}, since a fixed kernel only captures fine local details for near areas but broad global features for distant areas.

Deformable mechanisms~\cite{dai2017deformable, zhu2020deformable} have been proposed to address the fixed receptive field problem at the feature aggregation level.
However, relying on learnable offset parameters might be insufficient to consider camera perspective geometry.
To address this, we exploit a vanishing point (VP), which is a 2D point where parallel lines in 3D space appear to converge in the 2D image, as shown in \Fref{fig:overview}-(c).
Specifically, VP is a geometric cue that represents perspective geometry. It typically indicates distant areas in an image, especially in road environments where a single dominant VP is commonly observed.
Thus, by utilizing the VP, we can approximately distinguish whether a specific region in the images corresponds to near or distant areas.

In this work, we propose a framework named \texttt{VPOcc}, which utilizes the VP for 3D semantic occupancy prediction.
The proposed framework introduces three novel modules to mitigate the 2D-3D discrepancy of perspective geometry: VPZoomer, a VP-guided cross-attention (VPCA) module, and a spatial volume fusion (SVF) module.
Our contributions are summarized as follows:
\begin{itemize}
\item 
To the best of our knowledge, this is the first work to leverage the VP to improve neural 3D scene understanding.
\item 
We propose VPZoomer, the pixel-level module that counteracts the perspective effect by warping the image using a VP-based homography transformation.
\item 
We design the VP-guided cross-attention module, the feature-level solution that samples perspective-aware offset points based on the distance between the VP and the reference point.
\item
We introduce the spatial volume fusion module, which spatially integrates two feature volumes derived from the original and warped images to complement each other.
\item
By effectively addressing the 2D–3D discrepancy caused by perspective projection, our framework achieves superior performance on the SemanticKITTI~\cite{behley2019semantickitti} and SSCBench-KITTI360~\cite{li2023sscbench} datasets.
\end{itemize}

%

\section{Related Work}  \label{sec:related_work}
\subsection{3D semantic occupancy prediction}
3D semantic occupancy prediction, also known as 3D semantic scene completion (SSC), aims to simultaneously predict complete geometry and semantics from incomplete scene observations.
SSCNet~\cite{song2017semantic} is the first work to define the SSC task from a single RGB-D frame. 
Based on SSCNet, several studies utilize the 3D sensor data to improve SSC tasks~\cite{roldao2020lmscnet, mei2023ssc}.
Recently, camera-based 3D semantic occupancy prediction has gained significant attention in robot vision.
MonoScene~\cite{cao2022monoscene} is the first attempt to use only a single image.
It lifts 2D features into 3D by projecting voxels onto 2D feature maps, without using geometric priors.
Following MonoScene, several studies have been conducted to predict 3D semantic occupancy from a 2D image, focusing on effectively constructing 3D voxel feature volumes.
VoxFormer~\cite{li2023voxformer} constructs voxel queries based on depth estimated from images, and then lifts 2D feature maps into a 3D feature volume using cross-attention to place image features accurately.
OccFormer~\cite{zhang2023occformer} estimates a 3D voxel grid using the depth distribution of an image feature map and then employs a transformer-based decoder for completion.
Unlike previous works that primarily focus on constructing 3D feature volumes or designing decoders, we focus on leveraging appropriate image features for 3D understanding using a VP.

\subsection{Geometry as priors}
Camera geometry serves as a prior in various vision tasks.
In object detection, camera parameters are used to convert depth into point clouds~\cite{wang2019pseudo}, helping to overcome the limitations of 2D image representations.
Other works exploit the geometric structure of road environments for 3D object localization from images~\cite{choe2019segment2regress}.
In semantic segmentation, perspective-aware techniques are explored to improve the segmentation of distant regions~\cite{li2017foveanet}.
In depth estimation, methods consider the slanted angle of fisheye cameras to construct a ground-orthogonal feature space~\cite{lee2023slabins}.
In 3D semantic occupancy prediction, incorporating geometric priors is essential for accurately reconstructing a 3D scene from an image.
To this end, we exploit the VP as a novel geometric prior that reflects perspective geometry in the image.

\subsection{Image processing via zoom-in}
The lack of features in small regions of an image has been a fundamental issue in computer vision~\cite{liu2021survey}.
A common and intuitive solution is to zoom in on the image to enlarge small objects~\cite{zhang2019zoom}.
One approach is to resample the image, selectively enlarging regions of interest~\cite{thavamani2021fovea, thavamani2023learning}.
Another line of work enhances small object detection by leveraging both the upper and lower parts of the image plane, especially in road environments~\cite{ghosh2023learned}.
In contrast to these methods that typically zoom in on local areas, our VPZoomer enlarges distant areas and compresses near areas based on the VP, thereby restoring the amount of information (\ie, the number of pixels).

\begin{figure*}[t]
    \centering
    \includegraphics[width=0.99\linewidth]{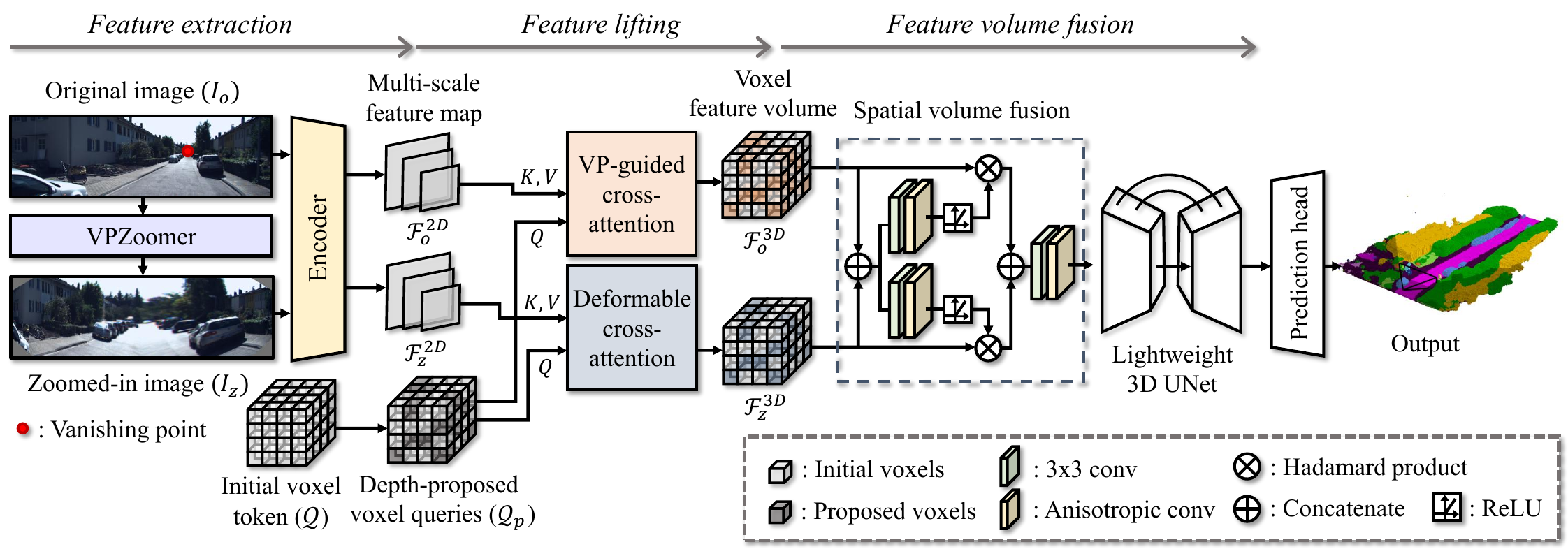}
    \caption{%
        \textbf{Overall architecture of \texttt{VPOcc}}. 
        In the feature extraction step, the zoomed-in image is generated using VPZoomer, and multi-scale feature maps $\mathcal{F}^{2D}_o$ and $\mathcal{F}^{2D}_z$ are extracted from $I_o$ and $I_z$.
        During the feature lifting, the depth-proposed voxel query $\mathcal{Q}_p$ is employed with the VP-guided cross-attention (VPCA) on $\mathcal{F}^{2D}_o$, and the deformable cross-attention on $\mathcal{F}^{2D}_z$ to construct the voxel feature volumes $\mathcal{F}^{3D}_o$ and $\mathcal{F}^{3D}_z$, respectively.
        In the feature volume fusion stage, both $\mathcal{F}^{3D}_o$ and $\mathcal{F}^{3D}_z$ are fused using a spatial volume fusion (SVF) module and refined via the 3D UNet-based decoder.
        Finally, the prediction head estimates the 3D semantic voxel map of the entire scene.}
    \label{fig:architecture}
    \vspace{-4mm}
\end{figure*}

\section{Method}  \label{sec:method}
We propose \texttt{VPOcc}, a camera-based 3D semantic occupancy prediction framework utilizing the VP.
As shown in \Fref{fig:architecture}, \texttt{VPOcc} takes a monocular image, the corresponding VP, and depth map as inputs\footnote{We utilize the pre-trained MobileStereoNet~\cite{shamsafar2022mobilestereonet} and NeurVPS~\cite{zhou2019neurvps} to estimate depth maps from stereo images and single dominant VPs from monocular images.} and estimates a 3D semantic voxel map.
Our framework comprises three components that incorporate VP-based perspective-aware modules: feature extraction, feature lifting, and feature volume fusion.
In the following sections, we describe the overall pipeline and provide details of the key modules in \texttt{VPOcc}.

\subsection{Overview}
\vspace{1mm}
\noindent \textbf{Feature extraction.} \
Given an input image $I_{o} \in \mathbb{R}^{H \times W \times 3}$ and its corresponding VP, we generate a zoomed-in image $I_{z} \in \mathbb{R}^{H \times W \times 3}$ using a proposed VP-based zoom-in module, VPZoomer.
It enlarges distant areas and compresses near areas, simultaneously creating a horizontally symmetric image around the image center by warping the original image based on the VP.
By utilizing both $I_{o}$ and $I_{z}$, we can leverage more balanced information across the scene by addressing the 2D-3D discrepancy at the pixel level.
Subsequently, we use an encoder\footnote{We use the pre-trained weights of MaskDINO~\cite{li2023mask}, following the approach of Symphonies~\cite{jiang2024symphonize}.} composed of ResNet-50~\cite{he2016deep} and the deformable transformer~\cite{zhu2020deformable} to create multi-scale feature maps $\mathcal{F}^{2D}_{o}$ and $\mathcal{F}^{2D}_{z}$ for each of $I_{o}$ and $I_{z}$.

\noindent \textbf{Feature lifting.} \ 
We construct a depth-proposed voxel query $\mathcal{Q}_{p}$, following Symphonies~\cite{jiang2024symphonize}.
First, we initialize voxel tokens $\mathcal{Q} \in \mathbb{R}^{ X \times Y \times Z \times C }$ with learnable embeddings.
Then, we back-project depth values from the image plane into 3D space using camera parameters, converting depth values into 3D points.
Voxel tokens that overlap with 3D points are employed as $\mathcal{Q}_{p}$.
Using $\mathcal{Q}_p$, we aggregate multi-scale image features via cross-attention to construct 3D voxel feature volumes.
In this process, our VP-guided cross-attention (VPCA) module, which samples points toward the VP, is applied on $\mathcal{F}^{2D}_{o}$ to generate $\mathcal{F}^{3D}_{o}\in \mathbb{R}^{X \times Y \times Z \times C}$.
With this perspective-aware 2D point sampling, we can aggregate 2D features with a granularity more suitable for 3D space.
Additionally, we utilize the general deformable cross-attention (DCA)~\cite{zhu2020deformable} to create $\mathcal{F}^{3D}_{z}\in \mathbb{R}^{X \times Y \times Z \times C}$ by aggregating 2D image features from $\mathcal{F}^{2D}_{z}$, which is extracted from the distorted image.
As a result, we obtain the voxel feature volumes $\mathcal{F}^{3D}_{o}$ and $\mathcal{F}^{3D}_{z}$ from the original and zoomed-in images, respectively.

\vspace{1mm}\noindent \textbf{Feature volume fusion.} \ 
$\mathcal{F}^{3D}_{o}$ and $\mathcal{F}^{3D}_{z}$ are fused through a spatial volume fusion (SVF) module and a lightweight 3D UNet-based decoder. 
By fusing them, each voxel aggregates a uniform amount of features and granularity from the image, regardless of its distance from the camera.
Lastly, the fused voxel feature volume is passed through a prediction head, which consists of 3D convolutions for upsampling to the target size, a 3D ASPP block~\cite{chen2017deeplab}, and a 3D convolution layer to generate the 3D semantic occupancy map.

\vspace{1mm}\noindent \textbf{Training objective.} \ 
Following MonoScene~\cite{cao2022monoscene}, we use scene-class affinity loss $\mathcal{L}_{\text{scal}}$ for class-wise metrics, with $\mathcal{L}_{\text{scal}}^{\text{sem}}$ and $\mathcal{L}_{\text{scal}}^{\text{geo}}$ for geometry and semantics.
We also employ a class-weighted cross-entropy loss $\mathcal{L}_{\text{ce}}$ for occupancy prediction.
The total loss is formulated as follows: 
\begin{equation}
    \mathcal{L} = \mathcal{L}_{\text{scal}}^{\text{geo}} + \mathcal{L}_{\text{scal}}^{\text{sem}} + \mathcal{L}_{\text{ce}}.
    \label{eqn:loss}
\end{equation}

\subsection{VPZoomer: VP-based image zoom-in} \label{ssec:vpzoomer}
\begin{figure}[t]
    \centering
    \includegraphics[width=0.98\linewidth]{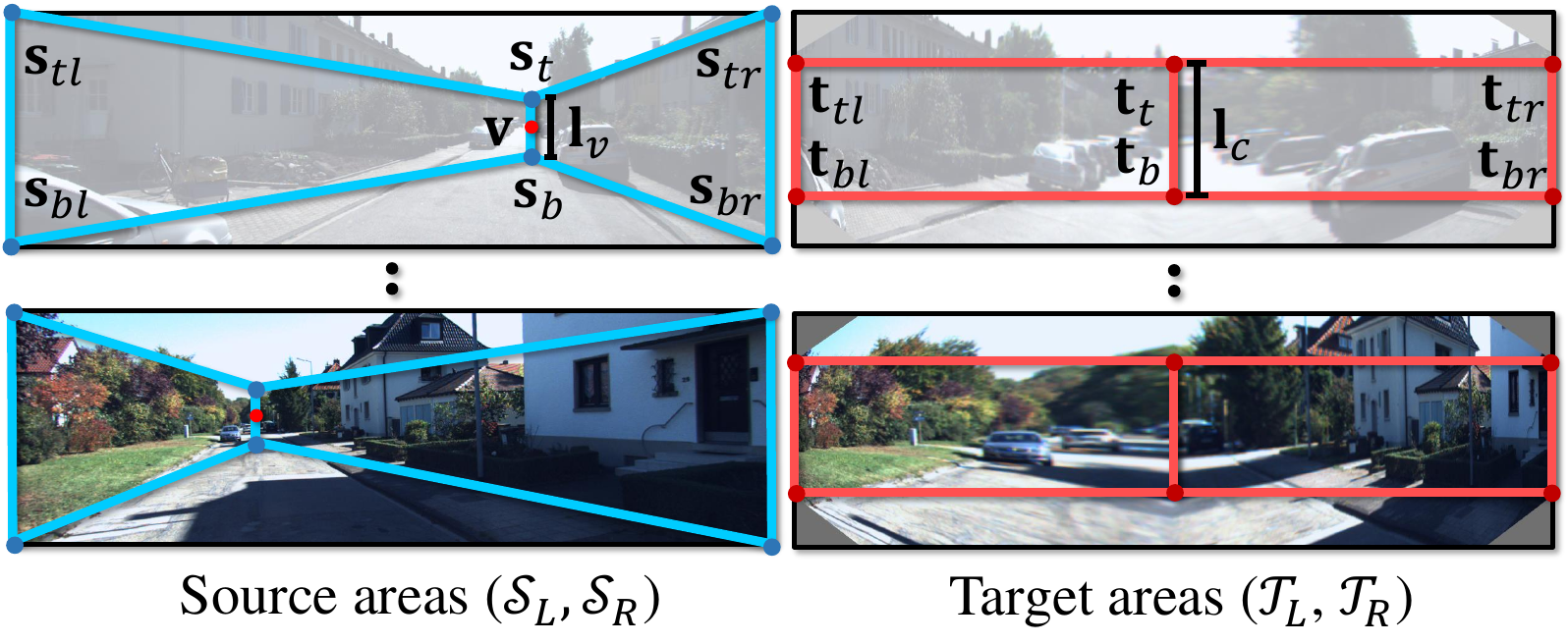}
    \caption{%
        \textbf{Illustration of VPZoomer}. 
        \emph{Left:} The original image $I_o$ with source areas ($\mathcal{S}_L$, $\mathcal{S}_R$) outlined in blue trapezoids. 
        \emph{Right:}~The zoomed-in image $I_z$ with target areas ($\mathcal{T}_L$, $\mathcal{T}_R$) outlined in red rectangles. 
    }
    \label{fig:vpzoomer}
\end{figure}

Due to camera perspective projection, 3D scenes projected onto 2D image planes result in an imbalanced pixel distribution at different distances from the camera.
To address this, our VPZoomer generates the zoomed-in image $I_{z}$ warped toward the VP.
To compensate for $I_o$, $I_z$ is designed to compress the near areas, enlarge the distant areas, and adjust the image to achieve horizontal symmetry (see \Fref{fig:vpzoomer}).
By utilizing both $I_o$ and $I_z$, we leverage a more balanced amount of pixels along the distance from the camera for the entire scene.

Given the original image $I_o$ and VP $\mathbf{v} = [v_x,v_y]^\top$, 
VPZoomer warps two source areas with trapezoidal shapes to target areas of rectangular shapes by 2D transformation.
It should be noted that we use trapezoids instead of VP-based triangles for source areas, as they reduce ambiguity and allow direct homography computation with their four vertices.

Specifically, we first need to define a shared vertical line to construct two trapezoids with a ribbon shape based on the VP. 
We define this shared vertical line $\mathbf{l}_v$ as a vertical line segment centered at VP $\mathbf{v}$ and with a length of $\alpha \cdot H$, where $\alpha$ is the scaling factor.
Here, we describe this line segment by two endpoints, top $\mathbf{s}_t$ and bottom $\mathbf{s}_b$:
\begin{equation}
    \mathbf{s}_t = [v_x, v_y - \frac{\alpha\cdot H}{2}]^\top, \
    \mathbf{s}_b = [v_x, v_y + \frac{\alpha\cdot H}{2}]^\top. 
\end{equation}
\noindent Based on the top $\mathbf{s}_t$ and the bottom $\mathbf{s}_b$ that define the vertical line $\mathbf{l}_v$, we construct two trapezoids, the left $\mathcal{S}_L$ and the right $\mathcal{S}_R$ trapezoid of source areas, each defined by a set of four vertices.
Let $\mathbf{s}_{tl}$, $\mathbf{s}_{tr}$, $\mathbf{s}_{bl}$, and $\mathbf{s}_{br}$\footnote{Subscript $l$, $r$, $t$ and $b$ indicate left, right, top and bottom, respectively.} be a set of corner vertices of the image. We can define two trapezoids as:
\begin{equation}
    \mathcal{S}_L = \{ \mathbf{s}_{tl}, \mathbf{s}_t, \mathbf{s}_b, \mathbf{s}_{bl} \}, \
    \mathcal{S}_R = \{ \mathbf{s}_t, \mathbf{s}_{tr}, \mathbf{s}_{br}, \mathbf{s}_b \}.
\label{eqn:trapezoid}
\end{equation}

For the target areas, we design them to compensate for the imbalance in the original image across depth and to achieve horizontal symmetry.
Thus, we define the target areas as two rectangles, {left $\mathcal{T}_L$ and right $\mathcal{T}_R$}, with a width of $W/2$. 
In other words, two rectangles share a vertical line $\mathbf{l}_c$ at the image center to have the balanced information density regardless of the two trapezoids of source areas.
Here, we also describe this line segment $\mathbf{l}_c$ by two endpoints, top $\mathbf{t}_t$ and bottom $\mathbf{t}_b$, and we stretch its length along the y-axis to increase pixel density at distant regions:
\begin{equation}
    \mathbf{t}_t = [\frac{W}{2}, \frac{\mathbf{s}_{ty}}{2} ]^\top, \
    \mathbf{t}_b = [\frac{W}{2}, \frac{H + \mathbf{s}_{by}}{2}]^\top. 
\label{eqn:dst_verticalline}
\end{equation}

By using two points $\mathbf{t}_t$ and $\mathbf{t}_b$, we can define two rectangles similarly to \Eref{eqn:trapezoid}:
\begin{equation}
    \mathcal{T}_L = \{ \mathbf{t}_{tl}, \mathbf{t}_t, \mathbf{t}_b, \mathbf{t}_{bl} \}, \
    \mathcal{T}_R = \{ \mathbf{t}_t, \mathbf{t}_{tr}, \mathbf{t}_{br}, \mathbf{t}_b \},
\end{equation}
where $\mathbf{t}_{tl}=[0, t_{ty}]^\top$, $\mathbf{t}_{tr}=[W, t_{ty}]^\top$, $\mathbf{t}_{bl}=[0, t_{by}]^\top$, and $\mathbf{t}_{br}=[W, t_{by}]^\top$. 
Note that $t_{ty}$ and $t_{by}$ represent the y-values of $\mathbf{t}_t$ and $\mathbf{t}_b$ in \Eref{eqn:dst_verticalline}, respectively.

After constructing correspondences between source and target (\ie, [$\mathcal{S}_L$, $\mathcal{T}_L$] and [$\mathcal{S}_R$, $\mathcal{T}_R$]), 
we then compute each homography transformation $\mathbf{H}_L$ and $\mathbf{H}_R$ by solving singular value decomposition (SVD)~\cite{hartley2003multiple}.
Lastly, the VP-guided zoom-in image $I_z$ is composited by the following process:
\begin{equation} 
\begin{gathered}
    I_z = \mathbf{M}\odot I_{l} + (1-\mathbf{M})\odot I_{r}, \\
    \textrm{where } I_r = \mathbf{H}_R(I_o), \ I_l =\mathbf{H}_L(I_o), 
\end{gathered}
\end{equation}
where $\mathbf{M}$ is a binary mask that activates the left-half area ($W/2$), $1-\mathbf{M}$ activates the remaining half area, $\odot$ indicates element-wise multiplication, and $\mathbf{H}_L(\cdot)$ and $\mathbf{H}_R(\cdot)$ represent warping functions. 

\begin{figure}[t]
    \centering
    \includegraphics[width=0.95\linewidth]{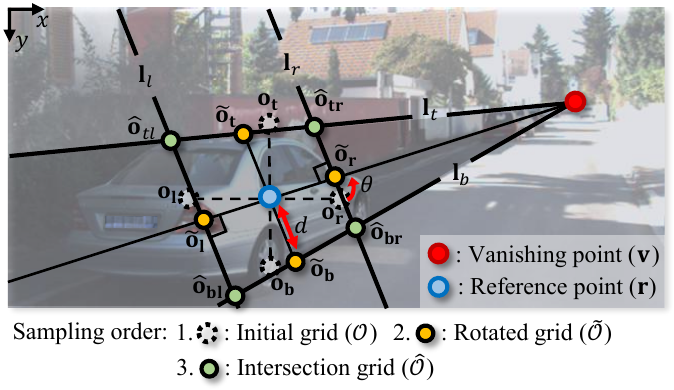}
    \caption{
        \textbf{VP-guided point sampling}.
        Following the sampling order, centered on the reference point $\mathbf{r}$, we first generate the initial grid $\mathcal{O}$ with the offset $d$ and rotate it by an angle $\theta$ to obtain $\tilde{\mathcal{O}}$. Next, we identify the intersection grid $\hat{\mathcal{O}}$ at the cross-point of lines from $\tilde{\mathcal{O}}$ and VP $\mathbf{v}$. As a result, a set of sampling points $\mathcal{P}$ is composed of $\{ \tilde{\mathcal{O}}, \hat{\mathcal{O}}, \mathbf{r} \}$. This process is further illustrated in the supplementary video.
    }
    \label{fig:vp_sampling}
\end{figure}

\subsection{VP-guided cross-attention} \label{ssec:vpca}
Recently, deformable cross-attention between 3D voxel queries and 2D image features has been commonly employed to lift 2D features into 3D~\cite{li2023voxformer, jiang2024symphonize}.
However, this approach only implicitly adjusts the offsets of sampling points without considering geometry between 2D and 3D (\ie, perspective projection).
To resolve this, we introduce the VP-guided cross-attention (VPCA) module, which samples points in a trapezoidal shape towards the VP. 
This scheme ensures perspective-aware feature aggregation, \corr{achieving more suitable feature granularity for 3D}.

\vspace{1mm}\noindent\textbf{VP-guided point sampling.} \
Given the VP $\mathbf{v}$ and a reference point $\mathbf{r}=[r_x, r_y]^\top$, which is the projected 2D position of the depth-proposed voxel queries $\mathcal{Q}_p$, VP-guided sampling aims to extract sample grid points $\mathcal{P}$ by considering the distance and the direction between $\mathbf{v}$ and $\mathbf{r}$.
Specifically, we first define cross-shaped initial grid points $\mathcal{O} = \{\mathbf{o}_l, \mathbf{o}_r, \mathbf{o}_t, \mathbf{o}_b\}$ with the offset along each axis (see \Fref{fig:vp_sampling}).
\rev{Next, we set the length of the offset $d$ as: 
\begin{equation}
    d = \frac{\| \mathbf{v} - \mathbf{r} \|_2}{2}. 
\label{eqn:offset}
\end{equation}}
Afterward, $d$ is bounded in the range of $[0, \beta\cdot c]$ to prevent an abnormal increase.
where $c$ indicates the scale parameter according to the resolution of feature $\mathcal{F}^{2D}_o$.
For example, $\mathbf{o}_l=[r_x-d, r_y]^\top$ and $\mathbf{o}_t=[r_x, r_y-d]^\top$ represent left grid and top grid points, respectively.
Then, using the angle formed by $\mathbf{v}-\mathbf{r}$ and the x-axis, we rotate $\mathcal{O}$ so that it is aligned with $\mathbf{v}$.

We denote these rotated initial grid points as $\tilde{\mathcal{O}}$.
Specifically, we compute the angle $\theta$ by $atan2(v_y - r_y, v_x - r_x)$ and then rotate each point $\mathbf{o} \in \mathcal{O}$:
\begin{equation}
    \tilde{\mathbf{o}} = \mathbf{R}_\theta \mathbf{o}, \ 
    \textrm{where } \mathbf{R}_\theta =
    \begin{bmatrix}
        \cos{\theta} & -\sin{\theta} \\
        \sin{\theta} & \cos{\theta}  \\
    \end{bmatrix}.   
\end{equation}
These rotated grid points $\tilde{\mathcal{O}} = \{ \tilde{\mathbf{o}}_l, \tilde{\mathbf{o}}_r, \tilde{\mathbf{o}}_t, \tilde{\mathbf{o}}_b \}$ correspond precisely to cross-shaped sample points that reflect a given $\mathbf{v}$ (see yellow points in \Fref{fig:vp_sampling}). 
Afterwards, we compute a line $\mathbf{l}_t$ that passes through $\mathbf{v}$ and $\tilde{\mathbf{o}}_t$ and the same goes for a line $\mathbf{l}_b$ by $\mathbf{v} \times \tilde{\mathbf{o}}_b$.  
From the direction obtained by subtracting $\tilde{\mathbf{o}}_t$ and $\tilde{\mathbf{o}}_b$, we can generate another line $\mathbf{l}_l$ that passes through $\tilde{\mathbf{o}}_l$, and the same goes for a line $\mathbf{l}_r$ for $\tilde{\mathbf{o}}_r$.
Then, by computing the intersection points between lines (\eg, $\hat{\mathbf{o}}_{tl} = \mathbf{l}_l \times \mathbf{l}_t$), we can estimate intersection grid points $\hat{\mathcal{O}} = \{ \hat{\mathbf{o}}_{tl}, \hat{\mathbf{o}}_{tr}, \hat{\mathbf{o}}_{bl}, \hat{\mathbf{o}}_{br} \}$ that construct a trapezoid shape toward the VP.
As a result, we can sample a set of nine points $\mathcal{P}=\{ \tilde{\mathcal{O}}, \hat{\mathcal{O}}, \mathbf{r} \}$ for each reference point $\mathbf{r}$.
We use $\mathcal{P}$ for the following VP-guided cross-attention.

\begin{table*}[t]
\caption{\textbf{Quantitative results on SemanticKITTI~\cite{behley2019semantickitti} test set.} $^\dagger$ denotes the results provided by \cite{roldao2020lmscnet}. $^\ast$ represents the reproduced results in \cite{huang2023tri, jiang2024symphonize}. The highest results among camera-based methods are denoted in \textbf{bold}, and the second-highest results are \underline{underlined}. The precision of decimals varies depending on the test values provided.}
    \centering
    \newcommand{\clsname}[2]{
        \rotatebox{90}{
            \hspace{-6pt}
            \textcolor{#2}{$\blacksquare$}
            \hspace{-6pt}
            \renewcommand\arraystretch{0.6}
            \begin{tabular}{l}
                #1                                      \\
                \hspace{-4pt} ~\tiny(\semkitfreq{#2}\%) \\
            \end{tabular}
        }}
    \renewcommand{\tabcolsep}{2pt}
    \renewcommand\arraystretch{1.1}
    \resizebox{0.98\textwidth}{!}
    {                          
        \begin{tabular}{l|c|c>{\columncolor{gray!15}}c|cccccccccccccccccccc}
            \toprule
            Method                               &
            Modality                             &
            \multicolumn{1}{c}{IoU}              &
            mIoU                                 &
            \clsname{road}{road}                 &
            \clsname{sidewalk}{sidewalk}         &
            \clsname{parking}{parking}           &
            \clsname{other-grnd.}{otherground}   &
            \clsname{building}{building}         &
            \clsname{car}{car}                   &
            \clsname{truck}{truck}               &
            \clsname{bicycle}{bicycle}           &
            \clsname{motorcycle}{motorcycle}     &
            \clsname{other-veh.}{othervehicle}   &
            \clsname{vegetation}{vegetation}     &
            \clsname{trunk}{trunk}               &
            \clsname{terrain}{terrain}           &
            \clsname{person}{person}             &
            \clsname{bicyclist}{bicyclist}       &
            \clsname{motorcyclist}{motorcyclist} &
            \clsname{fence}{fence}               &
            \clsname{pole}{pole}                 &
            \clsname{traf.-sign}{trafficsign}
            \\
            \midrule
            MonoScene$^\ast$~\cite{cao2022monoscene} & Camera & 34.16          & 11.08          & 54.70          & 27.10          & 24.80          & 5.70           & 14.40          & 18.80          & 3.30          & 0.50          & 0.70          & 4.40          & 14.90          & 2.40           & 19.50          & 1.00          & 1.40          & 0.40         & 11.10          & 3.30          & 2.10          \\
            TPVFormer~\cite{huang2023tri}   & Camera  & 34.25          & 11.26          & 55.10          & 27.20          & 27.40          & 6.50           & 14.80          & 19.20          & 3.70 & 1.00          & 0.50          & 2.30          & 13.90          & 2.60           & 20.40          & 1.10          & 2.40          & 0.30          & 11.00          & 2.90          & 1.50          \\
            
            VoxFormer-S~\cite{li2023voxformer}  & Camera  & 42.95 & 12.20          & 53.90          & 25.30          & 21.10          & 5.60           & 19.80          & 20.80          & 3.50          & 1.00          & 0.70          & 3.70          & 22.40          & 7.50           & 21.30          & 1.40          & 2.60 & 0.20          & 11.10          & 5.10          & 4.90          \\
            
            NDC-Scene~\cite{yao2023ndc} & Camera & 36.19 & 12.58 & 58.12 & 28.05 & 25.31 & 6.53 & 14.90 & 19.13 & 4.77 & 1.93 & 2.07 & \underline{6.69} & 17.94 & 3.49 & 25.01 & \textbf{3.44} & 2.77 & \underline{1.64} & 12.85 & 4.43 & 2.96 \\
            
            OccFormer~\cite{zhang2023occformer}   & Camera  & 34.53          & 12.32          & 55.90          & \underline{30.30} & \textbf{31.50} & 6.50           & 15.70          & 21.60          & 1.20          & 1.50          & 1.70          & 3.20          & 16.80          & 3.90           & 21.30          & 2.20          & 1.10          & 0.20          & 11.90          & 3.80          & 3.70          \\


            MonoOcc-S~\cite{zheng2024monoocc} & Camera  & - &  13.80 & 55.20 & 27.80 & 25.10 & 9.70 & 21.40 & 23.20 & \underline{5.20} & \underline{2.20} & 1.50 & 5.40 & 24.00 & 8.70 & 23.00 & 1.70 & 2.00 & 0.20 & 13.40 & 5.80 & 6.40 \\

            Symphonies~\cite{jiang2024symphonize}  & Camera & 42.19 & \underline{15.04} & \underline{58.40} & 29.30 & 26.90 & \textbf{11.70} & \underline{24.70} & 23.60 & 3.20 & \textbf{3.60} & \underline{2.60} & 5.60 & \underline{24.20} & \textbf{10.00} & 23.10 & \underline{3.20} & 1.90 & \textbf{2.00} & \underline{16.10} & \textbf{7.70} & \textbf{8.00} \\
            
            HASSC-S~\cite{wang2024not} & Camera & \underline{43.40} & 13.34 & 54.60 & 27.70 & 23.80 & 6.20 & 21.10 & 22.80 & 4.70 & 1.60 & 1.00 & 3.90 & 23.80 & 8.50 & 23.30 & 1.60 & \textbf{4.00} & 0.30 & 13.10 & 5.80 & 5.50 \\

            SGN-S~\cite{mei2024camera} & Camera & 41.88 & 14.01 & 57.80 & 29.20 & 27.70 & 5.20 & 23.90 & \underline{24.90} & 2.70 & 0.04 & 0.30 & 4.00 & \underline{24.20} & \textbf{10.00} & \underline{25.80} & 1.10 & 2.50 & 0.30 & 14.20 & \underline{7.40} & 4.40 \\

            \hline
            
            \texttt{VPOcc} (Ours)  & Camera & \textbf{44.58} &  \textbf{16.15} & \textbf{58.90} & \textbf{32.60} & \underline{30.18} & \underline{11.08} & \textbf{27.42} & \textbf{25.32} & \textbf{6.13} & 2.03 & \textbf{3.44} & \textbf{7.19} & \textbf{26.87} & \underline{9.32} & \textbf{27.04} & 2.31 & \underline{2.85} & 1.58 & \textbf{18.85} & 6.78 & \underline{6.92} \\
            
            \bottomrule
        \end{tabular}
       }

    \label{tab:sem_kitti_test}
    \vspace{-1mm}
\end{table*}

\vspace{1mm}\noindent \textbf{Feature aggregation by cross-attention.} \
VPCA aggregates image features at the sampled points $\mathcal{F}^{2D}_{o}(\mathbf{p}_s)$, where $\mathbf{p}_s \in \mathcal{P}$, using the depth-proposed voxel queries $\mathcal{Q}_p$ via a cross-attention mechanism, and generates the voxel feature volume $\mathcal{F}^{3D}_{o}$ as expressed as follows:
\begin{equation}
     \text{VPCA}(\mathcal{Q}_p, \mathbf{p}_s, \mathcal{F}^{2D}_o)=\displaystyle\sum_{i=1}^{N} \mathbf{A}_{i} \mathbf{W} \mathcal{F}^{2D}_{o}(\mathbf{p}^{i}_{s}), \label{eqn:vpca}
\end{equation}
where $N$ is the number of sampling points, $\mathbf{A}_{i}$ represents the attention weight, and $\mathbf{W}$ denotes the embedding projection weight.
For simplicity, we demonstrate only the single-head process for a single-scale 2D feature map.

\vspace{1mm}\noindent \textbf{Deformable cross-attention.} \
In addition to VPCA, we utilize a general deformable cross-attention (DCA)~\cite{zhu2020deformable} on $\mathcal{F}^{2D}_{z}$ to aggregate 2D image features from the distorted image by warping.
We project the depth-proposed voxel queries $\mathcal{Q}_p$ onto the 2D image plane to compose the reference points $\mathbf{p}_r$ and then compute the attention mechanism as follows:
\begin{equation}
     \text{DCA}(\mathcal{Q}_p, \mathbf{p}_r, \mathcal{F}^{2D}_z)=\displaystyle\sum_{i=1}^{N} \mathbf{A}_{i} \mathbf{W} \mathcal{F}^{2D}_{z}(\mathbf{p}_r+\Delta \mathbf{p}^{i}_{r}), \label{eqn:deformattn}
\end{equation} 
where the offset $\Delta \mathbf{p}^{i}_{r}$ indicates the predicted shift relative to $\mathbf{p}_r$.
As a result, we obtain the voxel feature volumes $\mathcal{F}^{3D}_{o}$ and $\mathcal{F}^{3D}_{z}$ from the original and zoomed-in images.

\subsection{Spatial volume fusion (SVF)} \label{sec:suff}
After the feature lifting process, we obtain two voxel feature volumes, $\mathcal{F}^{3D}_{z}$ and $\mathcal{F}^{3D}_{o}$, which address the perspective geometry at the pixel and feature levels, respectively.
To combine these two feature volumes, we introduce the spatial volume fusion module based on the gated fusion module of 3D-CVF~\cite{yoo20203d}.
Our module fuses $\mathcal{F}^{3D}_{z}$ and $\mathcal{F}^{3D}_{o}$, by locally and spatially integrating features from the original and the zoomed-in images.

As shown in \Fref{fig:architecture}, feature volume fusion begins by generating two attention masks from the voxel feature volumes $\mathcal{F}^{3D}_{o}$ and $\mathcal{F}^{3D}_{z}$ by convolutions.
The concatenated feature volume feeds forward through a 3$\times$3 convolution layer for local aggregation and an anisotropic convolution layer~\cite{li2020anisotropic} for spatial aggregation, which aggregates features anisotropically along each axis.
After that, each attention mask is generated via a ReLU activation function and is multiplied with each feature volume $\mathcal{F}^{3D}_{o}$ and $\mathcal{F}^{3D}_{z}$ to highlight important voxels.
The two attention-weighted volumes are concatenated and fused using the \joon{identically structured} local and spatial aggregation modules.
The fused feature volume is processed using the lightweight 3D UNet, which integrates a single 3D ASPP module as a bottleneck layer, inspired by MonoScene~\cite{cao2022monoscene}.

\begin{table*}[t]
\caption{\textbf{Quantitative results on SSCBench-KITTI360~\cite{li2023sscbench} test set.} We refer to the results in \cite{li2023sscbench, jiang2024symphonize}. 
The highest results among the camera-based methods are indicated in \textbf{bold}, and the second-highest results are \underline{underlined}.
}
    \centering
    \newcommand{\clsname}[2]{
        \rotatebox{90}{
            \hspace{-6pt}
            \textcolor{#2}{$\blacksquare$}
            \hspace{-6pt}
            \renewcommand\arraystretch{0.6}
            \begin{tabular}{l}
                #1                                       \\
                \hspace{-4pt} ~\tiny(\sscbkitfreq{#2}\%) \\
            \end{tabular}
        }}
    \newcommand{\empa}[1]{\textbf{#1}}
    \newcommand{\empb}[1]{\underline{#1}}
    \renewcommand{\tabcolsep}{2pt}
    \renewcommand\arraystretch{1.2}
    \resizebox{0.98\textwidth}{!}
    {
        \begin{tabular}{l|c|c>{\columncolor{gray!15}}c|rrrrrrrrrrrrrrrrrr}
            \toprule
            \multicolumn{1}{c|}{Method}             &
            Modality                                &
            IoU                                     &
            mIoU                                    &
            \clsname{car}{car}                      &
            \clsname{bicycle}{bicycle}              &
            \clsname{motorcycle}{motorcycle}        &
            \clsname{truck}{truck}                  &
            \clsname{other-veh.}{othervehicle}      &
            \clsname{person}{person}                &
            \clsname{road}{road}                    &
            \clsname{parking}{parking}              &
            \clsname{sidewalk}{sidewalk}           &
            \clsname{other-grnd.}{otherground}      &
            \clsname{building}{building}            &
            \clsname{fence}{fence}                 &
            \clsname{vegetation}{vegetation}        &
            \clsname{terrain}{terrain}              &
            \clsname{pole}{pole}                    &
            \clsname{traf.-sign}{trafficsign}       &
            \clsname{other-struct.}{otherstructure} &
            \clsname{other-obj.}{otherobject}
            \\
            \midrule
            MonoScene~\cite{cao2022monoscene}    & Camera        & 37.87               & 12.31        & 19.34        & 0.43        & 0.58        & 8.02         & 2.03         & 0.86        & 48.35        & 11.38        & 28.13        & 3.32        & 32.89        & 3.53        & 26.15        & 16.75        & 6.92         & 5.67        & 4.20         & 3.09         \\
            TPVFormer~\cite{huang2023tri}          & Camera    & 40.22       & 13.64        & 21.56        & 1.09        & 1.37        & 8.06         & 2.57         & 2.38        & 52.99        & 11.99        & 31.07        & 3.78        & 34.83        & 4.80        & 30.08        & 17.52        & 7.46         & 5.86        & 5.48         & 2.70         \\
            VoxFormer~\cite{li2023voxformer}       & Camera   & 38.76            & 11.91        & 17.84        & 1.16        & 0.89        & 4.56         & 2.06         & 1.63        & 47.01        & 9.67         & 27.21        & 2.89        & 31.18        & 4.97        & 28.99        & 14.69        & 6.51         & 6.92        & 3.79         & 2.43         \\
            OccFormer~\cite{zhang2023occformer}      & Camera     & 40.27          & 13.81        & 22.58        & 0.66        & 0.26        & 9.89         & 3.82         & 2.77        & 54.30        & 13.44        & 31.53        & 3.55        & 36.42 & 4.80        & 31.00        & 19.51 & 7.77         & 8.51        & 6.95         & 4.60         \\
            
            Symphonies~\cite{jiang2024symphonize}     & Camera & 44.12 & \underline{18.58} & \textbf{30.02} & 1.85 & \textbf{5.90} & \textbf{25.07} & \textbf{12.06} & \textbf{8.20} & 54.94 & 13.83 & 32.76  & \textbf{6.93} & 35.11  & \underline{8.58} & \textbf{38.33} & 11.52    & 14.01 & 9.57 & \textbf{14.44} & \textbf{11.28} \\
            
            SGN-S~\cite{mei2024camera} & Camera & \underline{46.22} & 17.71 & 28.20 & \underline{2.09} & 3.02 & 11.95 & 3.68 & 4.20 & \textbf{59.49} & \underline{14.50} & \underline{36.53} & 4.24 & \underline{39.79} & 7.14 & 36.61 & \textbf{23.10} & \textbf{14.86} & \underline{16.14} & 8.24 & 4.9 \\
            \hline

            \texttt{VPOcc} (Ours) & Camera & \textbf{46.39} &  \textbf{19.80} &  \underline{28.73} &  \textbf{2.93} & \underline{5.36} & \underline{21.71} & \underline{8.77} & \underline{6.42} & \underline{58.68} & \textbf{17.45} & \textbf{37.87} & \underline{5.37} & \textbf{42.62} & \textbf{8.91} & \underline{37.77} & \underline{22.44} & \underline{14.59} & \textbf{18.23} & \underline{10.48} & \underline{8.05} \\

            \bottomrule
        \end{tabular}
    }

    \label{tab:kitti_360_test}
    \vspace{-1mm}
\end{table*}
%

\begin{figure*}[t]
    \centering
    \includegraphics[width=0.95\linewidth]{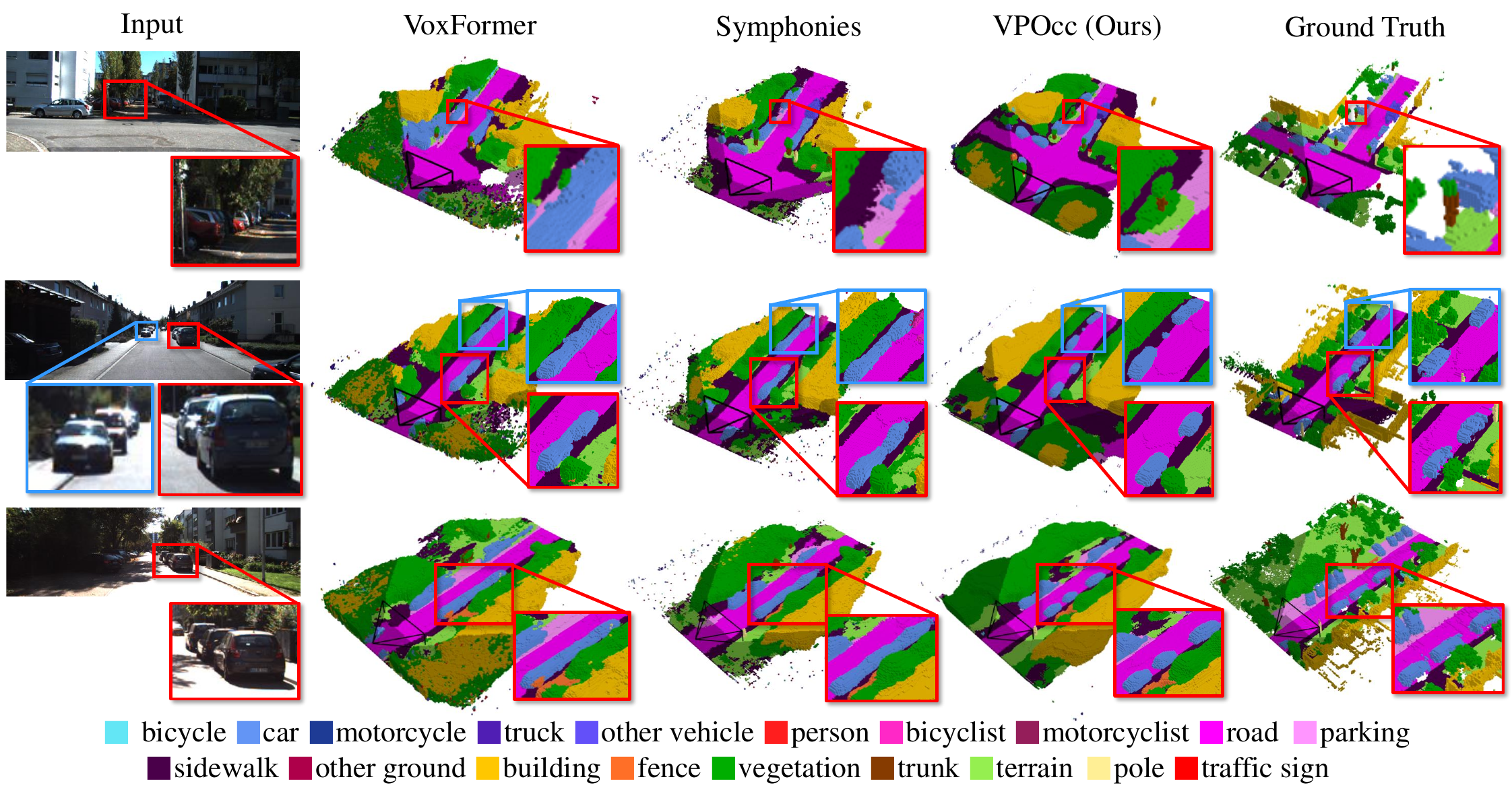}
    \vspace{-2mm}
    \caption{\textbf{Qualitative results on SemanticKITTI validation set}. Our \texttt{VPOcc} shows better performance \rev{along the road where usually VP is located. Additionally, boxed areas representing distant areas in the images demonstrate that our method also achieves superior performance in these distant areas.}}
    \label{fig:qualitative}
    \vspace{-4mm}
\end{figure*}

\section{Experiments}\label{sec:exp}
In this section, we evaluate our \texttt{VPOcc} and conduct ablation studies to validate the effectiveness of our framework. 
\Sref{subsec:performance} shows performance comparisons with the previous methods, and \Sref{subsec:ablations} validates the effectiveness of our modules through ablation studies.

\subsection{Implementation details} \label{sec:details}

\vspace{1mm}\noindent \textbf{Datasets and metrics.} \
We evaluate our method on the SemanticKITTI~\cite{behley2019semantickitti} and SSCBench-KITTI360~\cite{li2023sscbench} datasets, which contain 22 and 9 sequences with 20 and 19 classes, including the empty class.
Both datasets use $0.2 \text{m}$ voxel grids over a $51.2 \text{m} \times 51.2 \text{m} \times 6.4 \text{m}$ area, resulting in $256 \times 256 \times 32$ grids.
Performance is measured using IoU for binary completion and mIoU for semantic segmentation.

\vspace{1mm}\noindent \textbf{Training setup.} \
Experiments are conducted on 4 NVIDIA 3090 GPUs, training for 30 epochs with one batch per GPU.
We use the AdamW~\cite{loshchilov2017decoupled} optimizer with an initial learning rate of 2e-4 and a weight decay of 1e-4, reducing the learning rate to 0.1 times its original value after 25 epochs.
Data augmentation includes random horizontal flips.

\vspace{1mm}\noindent \textbf{Architecture details.} \ 
 \label{subsec:architecture}
We use RGB images, of which sizes are $1220\times 370$ for SemanticKITTI and $1408\times 376$ for SSCBench-KITTI360.
For the dimensions of $\mathcal{F}^{3D}$, denoted as $X \times Y \times Z$, we use $128 \times 128 \times 8$, which are subsequently upsampled to the scene resolution of $256\times 256\times 32$.
The hyperparameters are empirically determined.
$\alpha$ in VPZoomer is set to $0.2$.
In VPCA, we set the bound $\beta$ to $50$ and use scale factor $c \in \{ 1, 1.5, 2.0\}$ for feature resolutions $\mathcal{F}^{2D}_o \in \{ 1/4, 1/8, 1/16\}$.
We employ 3 layers, 9 sampling points, and 8 attention heads for each of VPCA and DCA.

\subsection{Performance comparison} \label{subsec:performance}

\noindent \textbf{Quantitative results.} \ 
As shown in \Tref{tab:sem_kitti_test} and \Tref{tab:kitti_360_test}, we compare existing 3D semantic occupancy prediction methods across different datasets: SemanticKITTI~\cite{behley2019semantickitti} and SSCBench-KITTI360~\cite{li2023sscbench}.
We outperform the previous methods, demonstrating improvements in both IoU ($+1.18$) and mIoU ($+1.11$) on the SemanticKITTI test set, as well as IoU ($+0.17$) and mIoU ($+1.22$) on the SSCBench-KITTI360 test set.
Unlike most of the existing methods (\eg, VoxFormer~\cite{li2023voxformer} and Symphonies~\cite{jiang2024symphonize}), which excel in only a single metric, our approach shows improvements in both IoU and mIoU metrics across all datasets.
We attribute this improvement to our attempt to address the perspective effect by utilizing the VP to enhance the comprehensive understanding of 3D scenes from 2D images, geometrically and semantically.

\vspace{1mm}\noindent \textbf{Qualitative results.}
In \Fref{fig:qualitative}, we provide a qualitative comparison result for the SemanticKITTI validation set.
Compared to VoxFormer and Symphonies, our \texttt{VPOcc} demonstrates sharp and distinguishable voxel shapes in both near and far areas.
By addressing the 2D-3D discrepancy problem caused by the camera perspective projection, our \texttt{VPOcc} effectively understands 3D scenes from 2D images.
For example, each boxed area shows that our approach enables the distinguishable and precise prediction of car shapes along the road by utilizing the VP.
Especially in the first column image, it is noticeable that only \texttt{VPOcc} constructs the tree and empty space in the distant areas.

\subsection{Ablation studies} \label{subsec:ablations}

We conduct ablation studies on SemanticKITTI~\cite{behley2019semantickitti} validation set to \corr{evaluate} our method.

\begin{table}[t!]
    \caption{\textbf{Ablation study for the proposed architecture.}}
    \centering
    \resizebox{\linewidth}{!}{%
        \def\arraystretch{1.3}
        \begin{tabular}{c|ccc|r>{\columncolor{gray!15}}c|cc}
            \toprule[0.3mm]
            Method                  & VPZoomer      & VPCA          & SVF      & IoU   & mIoU & Params (M)\\
            \midrule
            (1)                     & \checkmark    &               &           & 43.97 & 15.31 & 105.78 \\
            (2)                     & \checkmark    & \checkmark    &           & 44.18 & 15.79 & 105.72 \\
            \texttt{VPOcc} (Ours)   & \checkmark    & \checkmark    & \checkmark& \textbf{44.98} & \textbf{16.36} & 110.60 \\
            \bottomrule[0.3mm]
        \end{tabular}
    }
    \label{tab:arch}
    \vspace{2mm}
\end{table}

\begin{table}[t!]
    \caption{\textbf{Depth-wise performance evaluation.}}
    \centering
    \resizebox{\linewidth}{!}{%
        \def\arraystretch{1.3}
        \begin{tabular}{l|c>{\columncolor{gray!15}}c|c>{\columncolor{gray!10}}c|c>{\columncolor{gray!10}}c}
            \toprule[0.3mm]
            Range &
            \multicolumn{2}{c|}{$0m - 17m$} &
            \multicolumn{2}{c|}{$17m - 34m$} &
            \multicolumn{2}{c}{$34m - 51.2m$} \\
            \midrule
            Metric & IoU & mIoU & IoU & mIoU & IoU & mIoU \\
            \midrule
            MonoScene~\cite{cao2022monoscene} & 39.05 & 12.49 & 38.52 & 12.22 & 31.83 & 8.57 \\
            TPVFormer~\cite{huang2023tri} & 37.89 & 12.94 & 37.24 & 11.67 & 30.72 & 8.46 \\
            VoxFormer~\cite{li2023voxformer} & \textbf{48.60} & 12.95 & 46.83 & 13.85 & 35.85 & 9.54 \\
            OccFormer~\cite{zhang2023occformer} & 38.66 & 15.38 & 38.51 & 13.85 & 31.24 & 10.69 \\
            Symphonies~\cite{jiang2024symphonize} & 39.71 & 16.28 & 47.19 & 16.36 & 38.04 & 11.19 \\
            \hline
            \texttt{VPOcc} (Ours) & 44.97 & \textbf{17.12} & \textbf{49.66} & \textbf{18.11} & \textbf{39.74} & \textbf{13.22} \\
            \bottomrule[0.3mm]
        \end{tabular}
    }
    \label{tab:depth_wise}
\vspace{3mm}
\end{table}

\vspace{1mm}\noindent \textbf{Architecture composition.}
In this part, we validate the effectiveness of each module: VPZoomer, VPCA, and SVF.
\Tref{tab:arch} shows that each module of our framework contributes to the performance improvements in occupancy prediction and semantic segmentation.
First, \emph{Method (1)}, which uses VPZoomer with deformable cross-attention on $\mathcal{F}^{2D}_z$ for feature lifting and summation for fusion, demonstrates notable performance in this task.
This indicates that our architectural design, which handles the discrepancy in the number of pixels using VPZoomer, is effective.
In \emph{Method (2)}, by addressing the 2D-3D discrepancy at the feature level, we observe performance improvements in both IoU ($+0.21$) and mIoU ($+0.48$).
Lastly, by effectively fusing two feature volumes that address perspective geometry at both the pixel and feature levels, we achieve performance improvements in IoU ($+0.80$) and mIoU ($+0.57$).
Overall, this highlights the complementary design of \texttt{VPOcc}, where each module is structured to reinforce the others (\eg, VPZoomer–SVF and VPCA–DCA).

\vspace{1mm}\noindent\textbf{Depth-wise performance evaluation.}
To validate the effectiveness of our framework for each depth range, we conduct the performance evaluation by partitioning the scene into three areas based on depth.
As shown in \Tref{tab:depth_wise}, we can observe the performance improvements not only in far areas but also in near areas, especially in terms of mIoU.
This can be attributed to the inherent scale ambiguity in 3D scenes projected onto 2D images, caused by perspective projection and varying distances from the camera.
Our approach effectively addresses this challenge by employing VPZoomer to complement the pixel-level imbalance and VPCA to aggregate features with {adequate granularity.}
These components enable a more thorough understanding of 3D scenes from 2D images, leading {to performance improvements in overall areas.}

\begin{table}[t!]
    \caption{\textbf{Empirical study for fusion strategy.}}
    \centering
    \resizebox{0.85\linewidth}{!}{%
        \begin{tabular}{c|cccc}
            \toprule[0.3mm]
            Fusion & Sum & $1\times1$ Conv & Gated~\cite{yoo20203d} & SVF (Ours) \\
            \midrule
            IoU & 44.29 & 44.39 & 44.70 & \textbf{44.98} \\
            \rowcolor{gray!15}
            mIoU & 15.69 & 15.49 & 16.05 & \textbf{16.36} \\
            \bottomrule[0.3mm]
        \end{tabular}
    }
    \label{tab:feature_fusion}
\end{table}

\begin{table}[t]
\begin{minipage}[t]{0.45\linewidth}
    \caption{\textbf{Dependency on VP accuracy.}}
    \centering
    \resizebox{\linewidth}{!}{%
        \begin{tabular}{c|cc}
            \toprule[0.3mm]
            Method  & Na\"ive &  NeurVPS \\
            \midrule
            IoU  & 44.77  & \textbf{44.98} \\
            \rowcolor{gray!15}
            mIoU & 15.84  & \textbf{16.36} \\
            \bottomrule[0.3mm]
        \end{tabular}
    }\label{tab:vp}
\end{minipage}%
\hfill
\begin{minipage}[t]{0.54\linewidth}
    \caption{\textbf{Performance comparison with monocular depth.}}
    \centering
    \resizebox{\linewidth}{!}{%
        \begin{tabular}{c|c>{\columncolor{gray!15}}c}
            \toprule[0.3mm]
            Method  & IoU & mIoU \\
            \midrule
            Voxformer~\cite{li2023voxformer} & 38.68 &  10.67  \\
            Symphonies~\cite{jiang2024symphonize} & 38.37 & 12.20  \\
            SGN-S~\cite{mei2024camera} & 39.35 & 12.51 \\
            \hline
            \texttt{VPOcc} (Ours) & \textbf{40.92} & \textbf{13.15} \\
            \bottomrule[0.3mm]
        \end{tabular}
    }
    \label{tab:x_monocular}
\end{minipage}
\vspace{3mm}
\end{table}




\vspace{1mm}\noindent \textbf{Feature volume fusion.}
To effectively fuse the two different voxel feature volumes from the original and zoomed-in images, we evaluate various strategies: na\"ive summation (\ie, Sum), feature concatenation with $1\times1$ convolution (\ie, $1\times1$ Conv), adaptive gated feature fusion (\ie, Gated~\cite{yoo20203d}), and our spatial volume fusion module (\ie, SVF).
\Tref{tab:feature_fusion} shows that na\"ive summation is ineffective.
Per-voxel fusion via $1\times1$ Conv improves IoU ($+0.1$) but decreases mIoU ($-0.2$) due to missing spatial fusion.
Gated module, using local space attention, increases IoU ($+0.31$) and mIoU ($+0.56$) through adaptive local feature fusion.
Finally, our SVF module with anisotropic convolution~\cite{li2020anisotropic} aggregates local and broad features, achieving the highest performance in IoU ($+0.28$) and mIoU ($+0.31$).

\vspace{1mm}\noindent \textbf{Dependency on VP accuracy.} \
Since we use the estimated single dominant VP using pre-trained NeurVPS~\cite{zhou2019neurvps}, we evaluate the dependency of our framework on VP accuracy.
We compare the performance of using VP obtained from a less accurate na\"ive algorithm with that of VP from NeurVPS.
The na\"ive VP is obtained using Canny edge detection~\cite{canny1986computational}, the Hough transform~\cite{leavers1993hough}, and a voting grid cell.
As shown in \Tref{tab:vp}, the na\"ive VP leads to performance drop in both IoU ($-0.21$) and mIoU ($-0.52$).
Despite this, \texttt{VPOcc} remains robust demonstrating minimal reliance on VP accuracy.

\vspace{1mm}\noindent \textbf{Dependency on depth accuracy.}
We evaluate the performance of our method using less accurate monocular depth.
Specifically, we employ AdaBins~\cite{bhat2021adabins} to estimate monocular depth at a metric scale.
As shown in \Tref{tab:x_monocular}, our framework outperforms existing methods, achieving improvements in both IoU ($+1.57$) and mIoU ($+0.64$).
These results demonstrate the robustness of our framework even under inaccurate depth inputs.
\section{Conclusion}
In this work, we present \texttt{VPOcc}, a camera-based 3D semantic occupancy prediction framework that leverages the VP to mitigate the 2D-3D discrepancy problem  caused by perspective projection at both the pixel and feature levels.
First, at the pixel level, our VPZoomer module generates VP-based warped images to balance pixel density across varying distances from the camera.
Second, at the feature level, the VP-guided cross-attention (VPCA) module samples offset points toward VP to aggregate features with balanced granularity across distances, constructing a more representative 3D feature volume.
Finally, the spatial volume fusion (SVF) module resolves 2D-3D discrepancies at both the pixel and feature levels by effectively fusing the two feature volumes.
With the complementary architecture design across the modules, our \texttt{VPOcc} achieves improved performance in both IoU and mIoU across all datasets.

\vspace{1mm}\noindent \textbf{Limitations.} \
Since our method utilizes the pre-estimated VP, its performance partially depends on the accuracy of the VP.
We also assume a single dominant VP in road environments, which may limit robustness.
For future work, we plan to extend our approach to handle multiple VPs and improve generalization beyond the road scene.

\addtolength{\textheight}{-12cm}  

\bibliographystyle{IEEEtran}
\bibliography{IEEEabrv,edbib}

\begin{thebibliography}{10}
\providecommand{\url}[1]{#1}
\csname url@rmstyle\endcsname
\providecommand{\newblock}{\relax}
\providecommand{\bibinfo}[2]{#2}
\providecommand\BIBentrySTDinterwordspacing{\spaceskip=0pt\relax}
\providecommand\BIBentryALTinterwordstretchfactor{4}
\providecommand\BIBentryALTinterwordspacing{\spaceskip=\fontdimen2\font plus
\BIBentryALTinterwordstretchfactor\fontdimen3\font minus \fontdimen4\font\relax}
\providecommand\BIBforeignlanguage[2]{{%
\expandafter\ifx\csname l@#1\endcsname\relax
\typeout{** WARNING: IEEEtran.bst: No hyphenation pattern has been}%
\typeout{** loaded for the language `#1'. Using the pattern for}%
\typeout{** the default language instead.}%
\else
\language=\csname l@#1\endcsname
\fi
#2}}

\bibitem{wang2021learning}
L.~Wang, H.~Ye, Q.~Wang, Y.~Gao, C.~Xu, and F.~Gao, ``Learning-based 3d occupancy prediction for autonomous navigation in occluded environments,'' in \emph{IROS}, 2021.

\bibitem{crespo2020semantic}
J.~Crespo, J.~C. Castillo, O.~M. Mozos, and R.~Barber, ``Semantic information for robot navigation: A survey,'' \emph{Applied Sciences}, vol.~10, no.~2, p. 497, 2020.

\bibitem{cao2022monoscene}
A.-Q. Cao and R.~de~Charette, ``Monoscene: Monocular 3d semantic scene completion,'' in \emph{CVPR}, 2022.

\bibitem{li2023voxformer}
Y.~Li, Z.~Yu, C.~Choy, C.~Xiao, J.~M. Alvarez, S.~Fidler, C.~Feng, and A.~Anandkumar, ``Voxformer: Sparse voxel transformer for camera-based 3d semantic scene completion,'' in \emph{CVPR}, 2023.

\bibitem{huang2023tri}
Y.~Huang, W.~Zheng, Y.~Zhang, J.~Zhou, and J.~Lu, ``Tri-perspective view for vision-based 3d semantic occupancy prediction,'' in \emph{CVPR}, 2023.

\bibitem{yao2023ndc}
J.~Yao, C.~Li, K.~Sun, Y.~Cai, H.~Li, W.~Ouyang, and H.~Li, ``Ndc-scene: Boost monocular 3d semantic scene completion in normalized device coordinates space,'' in \emph{ICCV}, 2023.

\bibitem{zhang2023occformer}
Y.~Zhang, Z.~Zhu, and D.~Du, ``Occformer: Dual-path transformer for vision-based 3d semantic occupancy prediction,'' \emph{arXiv preprint arXiv:2304.05316}, 2023.

\bibitem{jiang2024symphonize}
H.~Jiang, T.~Cheng, N.~Gao, H.~Zhang, T.~Lin, W.~Liu, and X.~Wang, ``Symphonize 3d semantic scene completion with contextual instance queries,'' in \emph{CVPR}, 2024.

\bibitem{hartley2003multiple}
R.~Hartley and A.~Zisserman, \emph{Multiple view geometry in computer vision}.\hskip 1em plus 0.5em minus 0.4em\relax Cambridge university press, 2003.

\bibitem{dai2017deformable}
J.~Dai, H.~Qi, Y.~Xiong, Y.~Li, G.~Zhang, H.~Hu, and Y.~Wei, ``Deformable convolutional networks,'' in \emph{ICCV}, 2017.

\bibitem{zhu2020deformable}
X.~Zhu, W.~Su, L.~Lu, B.~Li, X.~Wang, and J.~Dai, ``Deformable detr: Deformable transformers for end-to-end object detection,'' \emph{arXiv preprint arXiv:2010.04159}, 2020.

\bibitem{behley2019semantickitti}
J.~Behley, M.~Garbade, A.~Milioto, J.~Quenzel, S.~Behnke, C.~Stachniss, and J.~Gall, ``{SemanticKITTI: A dataset for semantic scene understanding of lidar sequences},'' in \emph{ICCV}, 2019.

\bibitem{li2023sscbench}
Y.~Li, S.~Li, X.~Liu, M.~Gong, K.~Li, N.~Chen, Z.~Wang, Z.~Li, T.~Jiang, F.~Yu, \emph{et~al.}, ``Sscbench: A large-scale 3d semantic scene completion benchmark for autonomous driving,'' \emph{arXiv preprint arXiv:2306.09001}, 2023.

\bibitem{song2017semantic}
S.~Song, F.~Yu, A.~Zeng, A.~X. Chang, M.~Savva, and T.~Funkhouser, ``Semantic scene completion from a single depth image,'' in \emph{CVPR}, 2017.

\bibitem{roldao2020lmscnet}
L.~Roldao, R.~de~Charette, and A.~Verroust-Blondet, ``Lmscnet: Lightweight multiscale 3d semantic completion,'' in \emph{3DV}, 2020.

\bibitem{mei2023ssc}
J.~Mei, Y.~Yang, M.~Wang, T.~Huang, X.~Yang, and Y.~Liu, ``Ssc-rs: Elevate lidar semantic scene completion with representation separation and bev fusion,'' in \emph{IROS}, 2023.

\bibitem{wang2019pseudo}
Y.~Wang, W.-L. Chao, D.~Garg, B.~Hariharan, M.~Campbell, and K.~Q. Weinberger, ``Pseudo-lidar from visual depth estimation: Bridging the gap in 3d object detection for autonomous driving,'' in \emph{CVPR}, 2019.

\bibitem{choe2019segment2regress}
J.~Choe, K.~Joo, F.~Rameau, G.~Shim, and I.~S. Kweon, ``Segment2regress: Monocular 3d vehicle localization in two stages.'' in \emph{RSS}, 2019.

\bibitem{li2017foveanet}
X.~Li, Z.~Jie, W.~Wang, C.~Liu, J.~Yang, X.~Shen, Z.~Lin, Q.~Chen, S.~Yan, and J.~Feng, ``Foveanet: Perspective-aware urban scene parsing,'' in \emph{ICCV}, 2017.

\bibitem{lee2023slabins}
J.~Lee, G.~Cho, J.~Park, K.~Kim, S.~Lee, J.-H. Kim, S.-G. Jeong, and K.~Joo, ``{SlaBins: Fisheye Depth Estimation using Slanted Bins on Road Environments},'' in \emph{ICCV}, 2023.

\bibitem{liu2021survey}
Y.~Liu, P.~Sun, N.~Wergeles, and Y.~Shang, ``A survey and performance evaluation of deep learning methods for small object detection,'' \emph{Expert Systems with Applications}, vol. 172, p. 114602, 2021.

\bibitem{zhang2019zoom}
X.~Zhang, Q.~Chen, R.~Ng, and V.~Koltun, ``Zoom to learn, learn to zoom,'' in \emph{CVPR}, 2019.

\bibitem{thavamani2021fovea}
C.~Thavamani, M.~Li, N.~Cebron, and D.~Ramanan, ``Fovea: Foveated image magnification for autonomous navigation,'' in \emph{ICCV}, 2021.

\bibitem{thavamani2023learning}
C.~Thavamani, M.~Li, F.~Ferroni, and D.~Ramanan, ``Learning to zoom and unzoom,'' in \emph{CVPR}, 2023.

\bibitem{ghosh2023learned}
A.~Ghosh, N.~D. Reddy, C.~Mertz, and S.~G. Narasimhan, ``Learned two-plane perspective prior based image resampling for efficient object detection,'' in \emph{CVPR}, 2023.

\bibitem{shamsafar2022mobilestereonet}
F.~Shamsafar, S.~Woerz, R.~Rahim, and A.~Zell, ``Mobilestereonet: Towards lightweight deep networks for stereo matching,'' in \emph{WACV}, 2022.

\bibitem{zhou2019neurvps}
Y.~Zhou, H.~Qi, J.~Huang, and Y.~Ma, ``{NeurVPS}: Neural vanishing point scanning via conic convolution,'' in \emph{{NeurIPS}}, 2019.

\bibitem{li2023mask}
F.~Li, H.~Zhang, H.~Xu, S.~Liu, L.~Zhang, L.~M. Ni, and H.-Y. Shum, ``Mask dino: Towards a unified transformer-based framework for object detection and segmentation,'' in \emph{CVPR}, 2023.

\bibitem{he2016deep}
K.~He, X.~Zhang, S.~Ren, and J.~Sun, ``Deep residual learning for image recognition,'' in \emph{CVPR}, 2016.

\bibitem{chen2017deeplab}
L.-C. Chen, G.~Papandreou, I.~Kokkinos, K.~Murphy, and A.~L. Yuille, ``{Deeplab: Semantic image segmentation with deep convolutional nets, atrous convolution, and fully connected crfs},'' \emph{IEEE TPAMI}, vol.~40, no.~4, pp. 834--848, 2017.

\bibitem{zheng2024monoocc}
Y.~Zheng, X.~Li, P.~Li, Y.~Zheng, B.~Jin, C.~Zhong, X.~Long, H.~Zhao, and Q.~Zhang, ``Monoocc: Digging into monocular semantic occupancy prediction,'' \emph{arXiv preprint arXiv:2403.08766}, 2024.

\bibitem{wang2024not}
S.~Wang, J.~Yu, W.~Li, W.~Liu, X.~Liu, J.~Chen, and J.~Zhu, ``Not all voxels are equal: Hardness-aware semantic scene completion with self-distillation,'' in \emph{CVPR}, 2024.

\bibitem{mei2024camera}
J.~Mei, Y.~Yang, M.~Wang, J.~Zhu, J.~Ra, Y.~Ma, L.~Li, and Y.~Liu, ``Camera-based 3d semantic scene completion with sparse guidance network,'' \emph{IEEE TIP}, 2024.

\bibitem{yoo20203d}
J.~H. Yoo, Y.~Kim, J.~Kim, and J.~W. Choi, ``3d-cvf: Generating joint camera and lidar features using cross-view spatial feature fusion for 3d object detection,'' in \emph{ECCV}, 2020.

\bibitem{li2020anisotropic}
J.~Li, K.~Han, P.~Wang, Y.~Liu, and X.~Yuan, ``Anisotropic convolutional networks for 3d semantic scene completion,'' in \emph{CVPR}, 2020.

\bibitem{loshchilov2017decoupled}
I.~Loshchilov and F.~Hutter, ``Decoupled weight decay regularization,'' \emph{arXiv preprint arXiv:1711.05101}, 2017.

\bibitem{canny1986computational}
J.~Canny, ``A computational approach to edge detection,'' \emph{IEEE TPAMI}, no.~6, pp. 679--698, 1986.

\bibitem{leavers1993hough}
V.~Leavers, ``Which hough transform?'' \emph{CVGIP: Image understanding}, vol.~58, no.~2, pp. 250--264, 1993.

\bibitem{bhat2021adabins}
S.~F. Bhat, I.~Alhashim, and P.~Wonka, ``Adabins: Depth estimation using adaptive bins,'' in \emph{CVPR}, 2021, pp. 4009--4018.

\end{thebibliography}

\end{document}